\documentclass[runningheads]{llncs}
\usepackage[T1]{fontenc}
\usepackage{graphicx}
\usepackage{amsmath,graphicx,hyperref,amssymb,makecell,float,soul}
\usepackage{multirow}
\usepackage{booktabs}
\usepackage[misc]{ifsym}
\usepackage[table]{xcolor}
\usepackage[dvipsnames]{xcolor}
\usepackage{wrapfig}
\begin{document}
\title{Improving MLLM Training Efficiency via Stage-Aware Sparsity}
%
%
\author{Kean Shi\inst{1} \and
Liang Chen\inst{1} \and
Haozhe Zhao\inst{2} \and
Baobao Chang\inst{1}}
%
\authorrunning{Shi, Chen, Zhao, Chang et al.}
%
\institute{Peking University, Beijing, PRC \and
University of Illinois Urbana-Champaign, Champaign, IL, USA}
%
\maketitle              
\begin{abstract}

Multimodal Large Language Models (MLLMs) have demonstrated outstanding performance across a variety of domains. However, training MLLMs is often inefficient, as much of the computation is redundant due to the long input sequences from multimodal data and underutilized inter-layer operations. Notably, such redundancy is not static but varies across different stages of training. Building on this observation, we shift the focus to the training process itself and propose a training-efficient framework based on sparse representations, termed the Sparse Training Scheme (STS). Instead of applying a uniform sparsity strategy, STS adopts a stage-aware design that adapts to different sources of redundancy during training. Specifically, the framework consists of two complementary components: the Visual Token Compressor, which reduces the information load by compressing visual tokens during modality alignment, and the Layer Dynamic Skipper, which mitigates computational overhead by dynamically skipping unnecessary layers during instruction tuning. Our approach is broadly applicable to diverse MLLM architectures and has been extensively evaluated on multiple benchmarks, demonstrating its effectiveness and efficiency.

\end{abstract}

\section{Introduction}
\label{sec:intro}

Large Language Models (LLMs)~\cite{glm,gpt3,llama,qwen} have recently achieved a series of remarkable breakthroughs. The emergence of Multimodal Large Language Models (MLLMs)~\cite{blip2,internvl,qwenvl} such as Flamingo~\cite{flamingo}, GPT-4~\cite{gpt4}, and LLaVA~\cite{llava} signifies a clear shift of language models toward multimodal capabilities. However, the training and deployment of MLLMs face numerous technical challenges. Compared to LLMs, training MLLMs requires significantly more GPU memory and computation~\cite{wang2024visincontext}, as each sample contains not only text sequences but also hundreds of visual tokens, introducing substantial computational redundancy. For example, Qwen2.5-VL~\cite{qwen2.5vl} needs to process up to 16,384 visual tokens from images, and training Qwen2.5-VL-7B with a batch size of 16 typically requires at least 4 GPUs with 80 GB memory each, far exceeding the requirements of text-only LLMs~\cite{llama}.

Despite the strong capabilities of MLLMs, their training process contains substantial computational redundancy. Such redundancy arises from two primary sources. First, multimodal inputs introduce a large number of visual tokens, many of which contribute marginally to the final prediction, leading to \emph{data redundancy}. Second, large language models typically contain dozens of transformer layers, where not all layers are equally important throughout the entire training process, resulting in \emph{parameter redundancy}. While recent studies suggest that not all visual tokens or model parameters contribute equally during training~\cite{fastv}, existing efficiency methods usually address these two sources of redundancy independently, either by compressing visual tokens or by modifying model architectures. However, these approaches typically treat redundancy as static and overlook its evolving characteristics during different stages of training, which may lead to suboptimal efficiency-performance trade-offs.

In practice, the redundancy patterns of multimodal data and model parameters evolve differently across training stages. In the modality alignment stage, the language model is typically frozen and the computational bottleneck mainly comes from processing large numbers of visual tokens. In contrast, during instruction tuning, the language model parameters are actively updated and the computational overhead is dominated by deep transformer layers. This observation suggests that redundancy in MLLM training is inherently \emph{stage-dependent}, and thus improving training efficiency requires a \emph{stage-aware sparsity strategy} that adapts to different sources of redundancy at different stages.

Motivated by this insight, we propose a Sparse Training Scheme (STS) for MLLMs, a unified stage-aware sparse training framework that explicitly models the evolving redundancy patterns during training. Instead of applying a uniform sparsity strategy, STS dynamically adjusts the source of sparsity according to the dominant computational bottleneck at each stage. Specifically, STS can be instantiated through two complementary mechanisms. During modality alignment, a Visual Token Compressor (VTC) reduces redundant visual tokens before they are fed into the language model, thereby alleviating data redundancy. During instruction tuning, a Layer Dynamic Skipper (LDS) dynamically skips redundant decoder layers based on training progress and layer depth, reducing unnecessary parameter updates. By jointly addressing data and parameter redundancy across different training stages, STS provides a simple yet effective framework for improving the training efficiency of MLLMs.

\textbf{The main contributions of this work are as follows}:

\begin{itemize}
\item We analyze computational redundancy in MLLM training and reveal its \emph{stage-dependent} characteristics, showing that different sources of redundancy dominate at different training stages.

\item We propose a unified \emph{stage-aware sparse training framework} that adapts to evolving redundancy patterns, and provide an efficient instantiation that introduces token-level sparsity during modality alignment and layer-level sparsity during instruction tuning.

\item Extensive experiments on various benchmarks demonstrate both the effectiveness and practicality of our approach. On average, STS achieves a two-digit percentage reduction in training FLOPs while maintaining more than 96.9\% accuracy compared to baseline models.
\end{itemize}

\section{Related Work}
\label{sec:relatedworks}

\paragraph{Token-level Sparsity in Vision-Language Models.}
A large body of work focuses on reducing the computational overhead of vision-language models by compressing or pruning visual tokens~\cite{fastv,sparsevlm,llavolta}. 
Early studies observe that many visual tokens are redundant and contribute marginally to final predictions, motivating token pruning strategies. 
Representative approaches include FastV~\cite{fastv}, SparseVLM~\cite{sparsevlm}, and PruMerge~\cite{prumerge}, which selectively retain informative visual tokens based on attention or similarity scores. 
More recent methods explore adaptive token reduction, such as variation-aware pruning~\cite{v2drop} and dynamic strategies~\cite{dyrate}. 
These approaches effectively alleviate input redundancy, but primarily operate at the token level and treat redundancy as a static property.

\paragraph{Model-level Sparsity and Efficient Training.}
Another line of work improves efficiency by reducing computation in deep transformer architectures. 
Techniques such as stochastic depth~\cite{stochasticdepth} and LayerDrop~\cite{layerdrop} dynamically skip layers or reduce parameter updates during training and inference. 
In the context of large language models, staged training and sparsity-based optimization methods have also been explored~\cite{stagetrain4llm,notrainnogain}. 
While these approaches reduce parameter redundancy, they are typically designed for unimodal LLMs and do not explicitly consider multimodal interactions.

\begin{figure}[t]
\includegraphics[width=\textwidth]{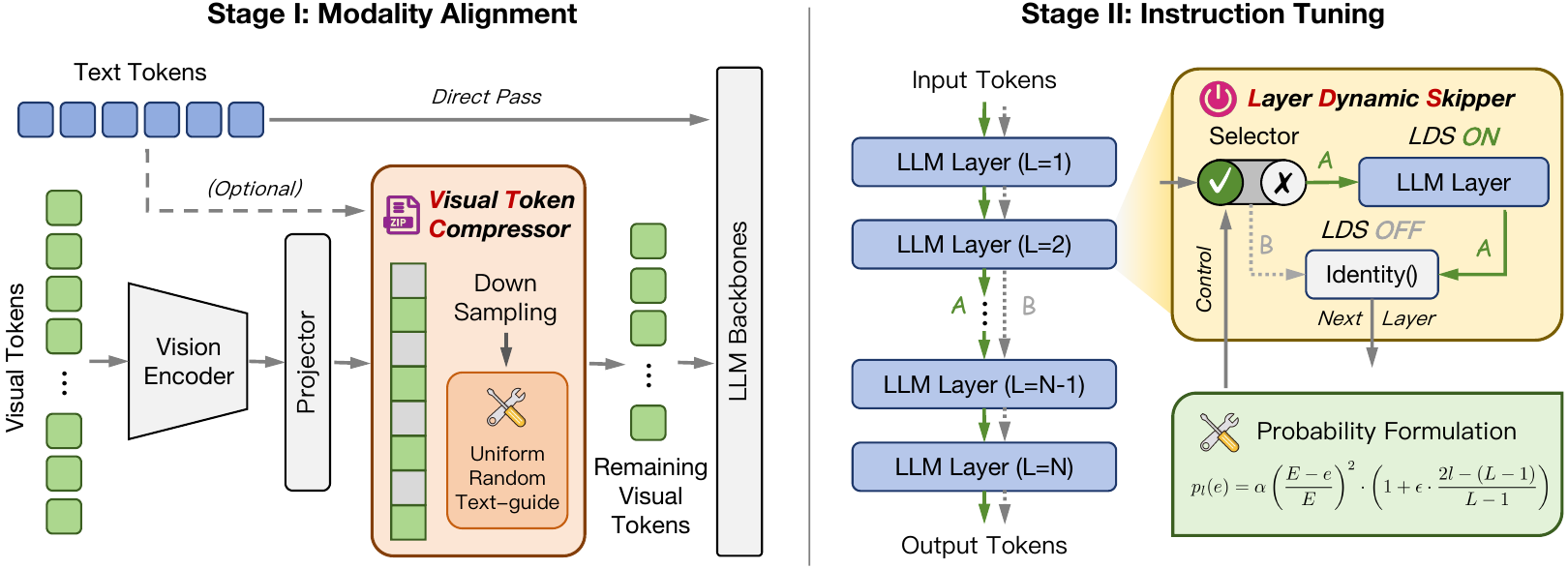}
\caption{
\textbf{Overview of the proposed stage-aware sparse training scheme (STS)}.
During modality alignment (Stage I), a visual token compressor reduces input redundancy by selecting informative visual tokens.
During instruction tuning (Stage II), a layer dynamic skipper adaptively skips redundant decoder layers based on training progress and layer depth.
The two components jointly address stage-dependent redundancy, improving training efficiency while preserving performance.
}
\label{fig:illustration}
\end{figure}

\section{Method}
\label{sec:format}

In this section, we first offer a brief introduction of naive training scheme in Sec~\ref{sec:preliminary}. As shown in Figure~\ref{fig:illustration}, We follow a stage-aware sparsity principle, where different sources of redundancy are addressed at different training stages. Our proposed STS includes two key components: VTC and LDS. They are detailed in Sec~\ref{sec:vtc} and Sec~\ref{sec:lds} respectively. Finally, we demonstrate the application of STS in the practical training of MLLMs in Section~\ref{sec:sts}.

\subsection{Preliminary}
\label{sec:preliminary}

The prevailing training scheme for MLLMs consists of two stages: \textbf{modality alignment} and \textbf{instruction tuning}.

In the \textit{modality alignment} stage, large-scale image-text pairs are used. A pretrained vision encoder $g(\cdot)$ extracts visual features from image tokens $\mathbf{X}_v$: $\mathbf{Z}_v = g(\mathbf{X}_v)$. $\mathbf{Z}_v$ are then projected into the language model’s embedding space via a learnable matrix $\mathbf{W}$ to obtain visual tokens: $\mathbf{T}_v = \mathbf{W} \mathbf{Z}_v$. The language model aligns visual and textual inputs by maximizing the likelihood of image-text pairs. Only $\mathbf{W}$ is updated during this stage.

In the \textit{instruction tuning} stage, dialogues sequence $(\mathbf{X}_q^1, \mathbf{X}_a^1, \dots, \mathbf{X}_q^T, \mathbf{X}_a^T)$ are used to train the model for instruction following. At each turn $t$, the instruction $\mathbf{X}_{\text{instruct}}^t$ is constructed from $\mathbf{X}_v$ and human query $\mathbf{X}_q^t$. The model autoregressively predicts the response $\mathbf{X}_a$ by maximizing:
\begin{equation}
    \label{eq:like}
    p(\mathbf{X}_a | \mathbf{X}_v, \mathbf{X}_{\text{instruct}}) = \prod_{i=1}^L p_\theta (x_i | \mathbf{X}_v, \mathbf{X}_{\text{instruct} <i}, \mathbf{X}_{a <i})
\end{equation}
In this stage, both the language model $f_{\phi}$ and $\mathbf{W}$ are updated. 

\subsection{Visual Token Compressor (VTC)}
\label{sec:vtc}

Due to the substantial redundancy in visual information~\cite{fastv,pddrop}, appropriately compressing the length of visual tokens is a simple yet effective approach to accelerate model training. We propose VTC, a selector can reduce the length of $\mathbf{T}_v$ and their hidden representations $\mathbf{H}_v$ within the LLM, while maximizing the likelihood of the target output $\mathbf{X_a}$ conditioned on the visual input and instruction prompt, i.e., $p(\mathbf{X}_a|\mathbf{X}_v,\mathbf{X}_{\mathrm{instruct}})$, as shown in Eq.~\ref{eq:like}.

The VTC can be described as function $\mathcal{C}$. In our setup, VTC is applied after the visual projector and before the LLM. Formally, given a sequence of visual tokens projected into a feature space $\mathbf{T}_v \in \mathbb{R}^{b \times n\times d}$, the compressor performs the following mapping:
\begin{equation}
    \label{eq:compressor1}
    \mathcal{C}: \mathbb{R}^{b \times n\times d} \rightarrow \mathbb{R}^{b \times n'\times d}, \ n'<n
\end{equation}
if $p$\% of visual tokens are expected to be retained, it can be calculated as,
\begin{equation}
    \label{eq:compressor2}
    \mathcal{C}(\mathbf{T}_v) = \mathbf{T}_v[:, \mathcal{I}_p, :]
\end{equation}
\noindent where $\mathcal{I}_p$ is the set of selected token indices.

It is worth noting that VTC supports multiple selection strategies by introducing different $\mathcal{I}_p$:

\noindent \textbf{Uniform Sampling.} Visual tokens are selected at fixed intervals according to a predefined stride, and unselected tokens are discarded. This strategy helps reduce local redundant details while preserving global semantic structures. The computation is given by:
\begin{equation}
    \label{eq:us}
    \mathcal{I}_p = \{0, s, 2s, \ldots\}, \quad s = \left\lfloor \frac{1}{p} \right\rfloor
\end{equation}

\noindent \textbf{Random Sampling.} A certain proportion of visual tokens are randomly discarded. This strategy improves the robustness of training by introducing stochasticity. The computation is given by:
\begin{equation}
    \mathcal{I}_p = \left\{ i \in \{0, 1, \ldots, n-1\} \mid z_i = 1 \right\}, \quad z_i \sim \text{Bernoulli}(p)
\end{equation}

\noindent \textbf{Instruction Guided.} Text tokens are generally considered to be important. Therefore, we leverage the cosine similarity between visual and text tokens to discard those irrelevant to the textual content. The selection for each visual token is computed as:
\begin{equation}
    \mathcal{I}_p = \operatorname{TopK}_{\lfloor p \cdot n \rfloor} \left( \operatorname{Mean}_{j} \left( \frac{\sum_{k=1}^d V_{ik} X_{jk}}{\sqrt{\sum_{k=1}^d V_{ik}^2} \sqrt{\sum_{k=1}^d X_{jk}^2}} \right) \right)
\end{equation}
where $V$ is the visual token and $X$ is the text token.

\subsection{Layer Dynamic Skipper (LDS)}
\label{sec:lds}

Since the major bottleneck in training lies in the computational overhead of large-scale LLMs, we propose LDS, a controller that dynamically adjusts the skipping probabilities of individual layers in the model, aiming to improve both training efficiency and model performance. Let \( \mathbf{H}_l \) denote the hidden state at the \( l \)-th layer and \( \mathcal{F}_l \) the encoding function of the \( l \)-th layer, LDS can be formalized as:
\begin{equation}
    \label{eq:LDS}
    \mathbf{H}_l = (1 - z_l) \cdot \mathcal{F}_l(\mathbf{H}_{l-1}) + z_l \cdot \mathbf{H}_{l-1}, \quad z_l \sim \text{Bernoulli}(p_l(e)) 
\end{equation}
where \( z_l \in \{0, 1\} \) controls whether the \( l \)-th layer is skipped; \( e \) denotes the current training step, and \( p_l(e) \) represents the skipping probability of the \( l \)-th layer at step \( e \).

Most MLLMs' training process exhibits a “coarse-to-fine” convergence trend, with parameters updating rapidly in the early stage and much more slowly later. This observation motivates the design of LDS. Since the early stage is dominated by coarse gradient updates, selectively skipping a portion of layers has little impact on overall convergence while greatly reducing computation. As gradients become more refined, skipping layers risks discarding subtle but important updates, so the skip rate should gradually decrease. To model this intuition, LDS adopts an step-based strategy: at the beginning, $\alpha$\% of the layers are skipped, and this rate smoothly decays to zero. Specifically, for the current step $e$, the probability of skipping is $\alpha \left( \tfrac{E - e}{E} \right)^2$, where $E$ is the total number of steps. The squared decay function allows a “fast-to-slow” drop in skip probability, aligning with the training loss trends.

Additionally, recent findings~\cite{wang2025exploring} suggest that shallow layers in LLMs are more effective in extracting data features than deeper layers. To incorporate this, LDS deploys a depth-aware adaptive skipping strategy. Specifically, for a given layer index $l$, the skip probability is defined as $1 + \epsilon \cdot \frac{2l - (L - 1)}{L - 1}$, where $L$ denotes the total number of layers, and $\epsilon$ is a hyperparameter that controls the variance of skip probabilities across different layers.

LDS integrates the above two strategies and controls layer skipping based on the probability derived from the following equation:
\begin{equation}
p_l(e) = \underbrace{\alpha \left( \frac{E - e}{E} \right)^2}_{\textit{step-based decay}} \cdot \underbrace{\left(1 + \epsilon \cdot \frac{2l - (L - 1)}{L - 1} \right)}_{\textit{depth-based adjustment}}
\label{eq:pl-e}
\end{equation}

\subsection{Sparse Training Scheme (STS)}
\label{sec:sts}

Building upon the two aforementioned components, we further design the STS applicable to most MLLMs. This design follows the stage-dependent redundancy observed in MLLM training. This scheme exploits both data and parameter redundancy in model training and applies sparsification according to different training stages.


\noindent \textbf{Modality Alignment.} In this stage, the LLM is typically frozen, and only the projector or vision encoder is updated. Therefore, VTC is applied to perform data sparsification, reducing the input length while preserving global visual information.  

\noindent \textbf{Instruction Tuning.} In this stage, the LLM parameters are updated, and the training data generally consist of carefully curated complex instructions. Thus, LDS is employed to alleviate the computational burden during training while retaining as much visual information as possible.

\begin{table*}[t]
    \centering
    \caption{\textbf{Main results of the proposed STS across various MLLMs.} We report performance on GQA, VQA$^{v2}$, SQA, POPE, and MME. Compared to the baseline, STS consistently reduces computational cost while maintaining comparable accuracy, with only marginal drops in benchmark scores. $^\dag$Activated parameters is counted only during the fine-tuning stage. $^\ddag$Average is calculated after converting MME to a 100-point scale.}
    \vspace{1em}
    
    \renewcommand{\arraystretch}{1.15} 
    \setlength{\tabcolsep}{5pt}        
    
    \resizebox{\textwidth}{!}{%
    \begin{tabular}{l c l c c ccccc c} 
    \toprule
    MLLM   &  LLM  &  Method  & \makecell{Activated\\Params.$^\dag$} & FLOPs & GQA  & VQA$^{v2}$ & SQA  & POPE & MME & Average$^\ddag$  \\ 
    \midrule 
    
    \multirow{2}{*}{LLaVA}  &  \multirow{2}{*}{Vicuna} & Baseline & 0.32B &  -  & 63.0 & 79.1  & 68.4 & 86.4 & 1476.9 & 74.1        \\  
      &  & \textbf{STS}  &  0.27B   &  82.2\%  & 59.7 &  76.5  & 70.2 & 86.4 & 1451.5 & 73.1\color{red}{$_{98.7\%}$}      \\ 
    \midrule 
    
    \multirow{4}{*}{Mipha} & \multirow{2}{*}{Phi-1.5} & Baseline & 1.30B & -  & 59.8 &  76.2  & 69.9 & 86.0 & 1394.0 & 72.3            \\
     & & \textbf{STS}  &  1.08B  &  83.5\%  &  59.2 &  74.8 & 68.9 & 86.7 & 1393.8 & 72.1\color{red}{$_{99.7\%}$}                \\  \cmidrule(lr){2-11} 
    
     &  \multirow{2}{*}{Phi-2}  & Baseline &  2.70B  & - & 59.0 & 76.3  & 57.5 & 87.3 & 1189.1 & 67.9             \\
     &  & \textbf{STS}      & 2.24B & 83.8\%  & 58.1 &  75.8  & 55.7 & 87.6 & 1161.2 & 66.9\color{red}{$_{98.5\%}$}                \\ 
    \midrule 
    
    \multirow{4}{*}{MoE-LLaVA} & \multirow{2}{*}{Qwen} & Baseline & 1.80B  & - & 61.5 & 76.2  & 63.1 & 87.0 & 1291.6 & 70.5                \\
     & & \textbf{STS}      & 1.49B  & 83.9\% & 57.9 &  74.0 & 62.1 & 86.2 & 1230.1 & 68.3\color{red}{$_{96.9\%}$}                \\  \cmidrule(lr){2-11} 
    
     & \multirow{2}{*}{StableLM} & Baseline &  1.60B & - & 60.3 & 76.7  & 62.6 & 85.7 & 1318.2 & 70.2                \\
     & & \textbf{STS}      &  1.33B  &  83.9\% & 57.0 &  74.5 & 63.4 & 86.0 & 1223.2 & 68.4\color{red}{$_{97.4\%}$}                \\ 
    \bottomrule
    \end{tabular}%
    } 
    \label{tab:main}
\end{table*}

\section{Experiment}
\label{sec:exp}

In this section, we first describe the implementation details of our experiments in Sec~\ref{sec:implement}, followed by discussing the main results across various MLLMs in Sec~\ref{sec:main}. We also illustrate the ablation studies in Sec.~\ref{sec:ablation} and computational efficiency analysis in Sec.~\ref{sec:efficiency}.

\subsection{Implementation Details}
\label{sec:implement}

We evaluate our STS on LLaVA~\cite{llava}, Mipha~\cite{mipha}, and MoE-LLaVA~\cite{moellava}, adhering strictly to the hyperparameter settings provided in their original training scripts. During modality alignment, we apply VTC to compress visual tokens, while in instruction tuning we enable LDS to dynamically skip layers of the LLM. After training every model for 1 epoch, we adopt several benchmarks specifically designed for MLLM evaluation, including GQA~\cite{gqa}, VQA$^{v2}$~\cite{VQAv2}, SQA~\cite{sqa}, MME~\cite{mme} and POPE~\cite{pope}.

\subsection{Main Results}
\label{sec:main}

To evaluate the effectiveness and efficiency of our proposed STS method, we conduct experiments on five representative MLLMs. We adopt a collection of benchmarks that together provide a comprehensive view of model capability: (i) perception tasks such as GQA, VQA$^{v2}$, and SQA, which evaluate reasoning and visual understanding; (ii) robustness assessment via POPE, which measures resistance to hallucination; and (iii) comprehensive evaluation with MME, which covers multiple vision–language abilities.

As shown in Table~\ref{tab:main}, STS consistently reduces the number of activated parameters and input length across all models. This reduction translates into an average FLOPs saving of over 15\%, confirming the efficiency benefits of our approach. Importantly, these savings come with only marginal performance degradation. On average, accuracy is preserved at 96.9–99.7\% of the baseline, with the largest observed drop being 2.2 points. The trend remains consistent across model families and scales, all these architectures show similar improvements in efficiency with only minor trade-offs in accuracy.

Breaking down the results by benchmarks, we observe that STS maintains stable performance on POPE and MME, confirming that robustness and broad vision–language skills are largely unaffected by sparsification. On perception-oriented tasks, such as GQA and VQA$^{v2}$, performance drops are slightly more noticeable, though still around 2–3 points, which highlights a modest sensitivity of fine-grained reasoning tasks to parameter reduction. Interestingly, in some cases (e.g., SQA for LLaVA), STS even outperforms the baseline, suggesting that selective computation may act as a form of implicit regularization.

\begin{table}[t]
    \centering
    \caption{\textbf{Ablation study on different pre-training acceleration strategies.} $^\dag$CSR: Commonsense Reasoning. \sethlcolor{green!10}\hl{Green row} denotes the default strategy of STS.}
    \vspace{1em}
    \renewcommand{\arraystretch}{1.1}
    \setlength{\tabcolsep}{10pt}
    \begin{tabular}{lcccc}
    \toprule
    Strategy  & Hallucination & Count & OCR   & CSR$^\dag$ \\
    \midrule
    Random  & 86.3 & 150.0 & 100.0 & 121.4 \\
    Instruction  & 86.4 & 155.0 & 92.5  & 126.4 \\
    \rowcolor{green!10} Uniform & 86.4 & 155.0 & 100.0 & 125.0 \\
    \bottomrule
    \end{tabular}
    \label{tab:ab1}
\end{table}

\begin{table}[t]
    \centering
    \caption{\textbf{Ablation study on layer skipping hyperparameters.} We compare different values of $\alpha$ and $\epsilon$ in Eq.~\ref{eq:pl-e}. \sethlcolor{green!10}\hl{Green row} denotes the default hyperparameters used in STS.}
    \vspace{1em}
    \renewcommand{\arraystretch}{1.1}
    \setlength{\tabcolsep}{10pt}
    \begin{tabular}{cccccc}
    \toprule
    \multicolumn{2}{c}{Hyper-Parameter} & Hallucination & Count & OCR  & CSR \\ \midrule
    \multirow{2}{*}{$\alpha=0.5$} & $\epsilon=0.3$ & 86.5 & 138.3 & 95.0 & 114.3 \\
                                  & $\epsilon=0.7$ & 86.5 & 153.3 & 92.5 & 121.4 \\ \midrule
    \multirow{2}{*}{$\epsilon=0.5$} & $\alpha=0.3$ & 86.6 & 163.3 & 117.5 & 127.9 \\
                                  & $\alpha=0.7$ & 86.5 & 135.0 & 80.0 & 123.6 \\ \midrule
    \rowcolor{green!10} $\alpha=0.5$ & $\epsilon=0.5$ & 86.4 & 155.0 & 100.0 & 125.0 \\ \bottomrule                            
    \end{tabular}
    \label{tab:ab2}
\end{table}

\subsection{Ablation Studies}
\label{sec:ablation}

\textbf{VTC Ablation.}
To better understand the variants of VTC, we conduct an ablation study comparing Random, Instruction-based, and Uniform sampling. We want to determine how different strategies influence downstream performance across four aspects: Hallucination, Counting, OCR, and Commonsense Reasoning. As summarized in Table~\ref{tab:ab1}, Random selection produces inconsistent outcomes, highlighting its instability across tasks. Instruction-based sampling provides a moderate boost in CSR but significantly reduces OCR accuracy, indicating a bias toward reasoning tasks at the expense of perception ability. By contrast, the Uniform strategy achieves more balanced performance across all four metrics, avoiding large trade-offs between different task categories. These results suggest that a uniform allocation of training effort across task types is more robust and reliable, so we adopt Uniform sampling as our default strategy.

\noindent \textbf{Hyper-parameter Study.}
We further examine the sensitivity of LDS to its two key hyperparameters, $\alpha$ and $\epsilon$ in Eq.~\ref{eq:pl-e}, to explore how different settings affect the efficiency–performance trade-off. As shown in Table~\ref{tab:ab2}, smaller $\alpha$ emphasizes deeper training, yielding higher accuracy but increasing cost, while larger $\alpha$ improves efficiency at the expense of task performance. Similarly, $\epsilon$ redistributes the skipping probability across layers: $\epsilon{=}0.7$ favors deeper layer skipping and benefits CSR but compromises OCR, whereas $\epsilon{=}0.3$ spreads skipping more evenly but weakens shallow-layer perception, reducing overall performance. Taken together, these results illustrate that extreme values of either parameter tend to bias the model toward certain abilities. The configuration $\alpha{=}0.5, \epsilon{=}0.5$ provides the best balance across tasks, so we adopt it as the default setting for LDS.

\begin{figure}[t]
    \centering
    \includegraphics[width=0.8\linewidth]{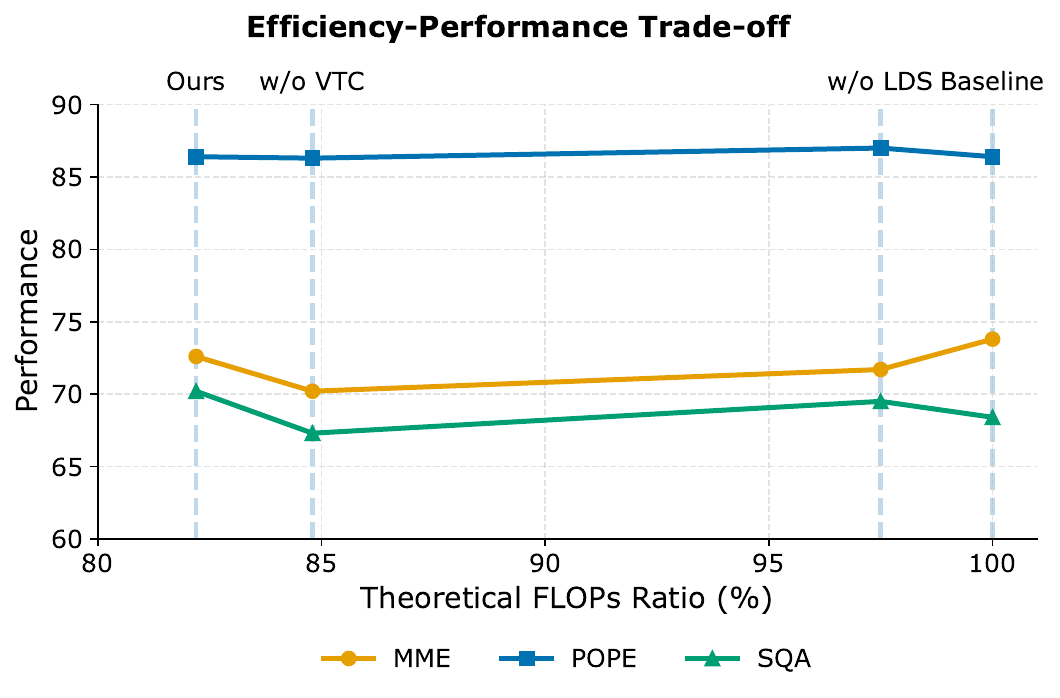}
    \vspace{-2em}
    \caption{\textbf{Efficiency–performance trade-off across different acceleration strategies.} We compare baseline, w/o VTC, w/o LDS, and our STS on three benchmarks including MME, POPE, and SQA. The x-axis denotes the ratio of theoretical FLOPs relative to the baseline, while the y-axis reports task-specific accuracy. The MME score has been divided by 2000.}
    \label{fig:trade-off}
\end{figure}

\subsection{Computational Efficiency}
\label{sec:efficiency}

We further evaluate the computational efficiency of STS by analyzing FLOPs reduction and performance retention across MME, POPE, and SQA. As shown in Figure~\ref{fig:trade-off}, the joint framework that integrates VTC and LDS achieves the most substantial reduction in FLOPs, down to 82.2\%, while maintaining accuracy comparable to the baseline across all tasks. In contrast, using either component in isolation shifts the balance unfavorably. VTC alone yields only marginal computational savings, whereas LDS alone achieves greater cost reduction but at the expense of accuracy. The results demonstrate that the two components address complementary aspects of efficiency, and highlight that STS achieves a more favorable trade-off between efficiency and accuracy, establishing its effectiveness as a general training scheme for MLLMs.

\section{Conclusion}
\label{conclusion}

We presented STS, a stage-aware framework for improving the training efficiency of multimodal large language models. Instead of applying a uniform sparsity strategy, STS adapts to stage-dependent redundancy by compressing redundant visual inputs and skipping unnecessary decoder layers through VTC and LDS, respectively. Experiments across multiple benchmarks show that STS achieves faster training with minimal performance loss. Overall, STS provides a practical and scalable solution for efficient MLLM optimization.

%
%
%
\bibliographystyle{splncs04}
\bibliography{mybibliography}
\end{document}